# Corpus multilingues pour l'étude de nouveaux concepts en SHS :

*le cas de « l'innovation non technologique »*

- Kyriakoglou Revekka (1), MCF, https://orcid.org/0009-0007-8626-8401. Mail : kyriakoglou@up8.edu.
- Pappa Anna (1), PU, https://orcid.org/0000-0003-2447-4078. Mail : ap@up8.edu.
(1) LIASD – Université Paris 8 – UR 4383

- Informatique (Traitement Automatique des Langues)
- Sciences humaines et sociales, Ingénierie linguistique, Corpus multilingues
- Construction et documentation d'un corpus / pipeline de collecte / dataset dérivé
- Corpus multilingue, web scraping, extraction contextes, annotation catégorielle, concepts émergents, SHS, apprentissage automatique
- Multilingual corpus, web scraping, context extraction, categorical annotation, emerging concepts, HSS, machine learning

- Cet article présente une méthodologie hybride pour la création d'un corpus multilingue destiné à faire émerger des termes associés à des concepts nouveaux en sciences humaines et sociales (SHS), en prenant pour cas d'étude « l'innovation non technologique ». Le corpus repose sur deux sources textuelles complémentaires : (1) des contenus extraits automatiquement des sites web d'entreprises, nettoyés pour le français et l'anglais, et (2) des rapports annuels collectés et filtrés automatiquement selon des critères documentaires (année, format, duplication). Le pipeline intègre la détection automatique des langues, le filtrage des contenus non pertinents, l'extraction de segments pertinents et l'enrichissement en métadonnées structurelles.

À partir de ce corpus initial, un dataset dérivé en anglais est construit pour des usages en apprentissage automatique. Il regroupe, pour chaque occurrence d'un terme du lexique expert, un bloc contextuel de cinq phrases (deux précédant et deux suivant la phrase contenant le terme). Chaque occurrence est annotée avec la catégorie thématique associée au terme, permettant de structurer les données pour des tâches de l'apprentissage automatique.

Cette approche produit une ressource reproductible, extensible et adaptée à l'analyse de la variabilité lexicale autour de concepts nouveaux, ainsi qu'à la création de jeux de données destinés aux applications de traitement automatique des langues.

- This article presents a hybrid methodology for building a multilingual corpus designed to support the study of emerging concepts in the humanities and social sciences (HSS), illustrated here through the case of "non-technological innovation". The corpus relies on two complementary sources: (1) textual content automatically extracted from company websites, cleaned for French and English, and (2) annual reports collected and automatically filtered according to documentary criteria (year, format, duplication). The processing pipeline includes automatic language detection, filtering of non-relevant content, extraction of relevant segments, and enrichment with structural metadata. From this initial corpus, a derived dataset in English is created for machine learning purposes. For each occurrence of a term from the expert lexicon, a contextual block of five sentences is extracted (two preceding and two following the sentence containing the term). Each occurrence is annotated with the thematic category associated with the term, enabling the construction of data suitable for supervised classification tasks.

This approach results in a reproducible and extensible resource, suitable both for analyzing lexical variability around emerging concepts and for generating datasets dedicated to natural language processing applications.

# 1. INTRODUCTION

L'utilisation de corpus spécifiques constitue un enjeu central pour l'analyse des concepts émergents en sciences humaines et sociales (SHS), en particulier lorsqu'il s'agit des notions encore peu couvertes ou insuffisamment représentées dans les ressources lexicales existantes. Plusieurs structuration conceptuelles dérivées de *Princeton WordNet* [1] ont été proposées, comme *WordNet Domains* qui ajoute une hiérarchie thématique aux *synsets* [1] puis son extention multilingue révisée [2] D'autres ressources de large couverture, telles que *BabelNet, qui combine WordNet et Wikipedia* [3], *Open Multilingual WordNet* [4] et *ConceptNet* [5] offrent des représentations conceptuelles utilisées pour l'analyse textuelle. Toutefois, ces ressources, centrées sur des concepts stabilisés et encyclopédiques, se révèlent moins adaptées à l'étude de notions émergentes ou interdisciplinaires, qui nécessitent des corpus thématiques contextualisés et dynamiques.

L'exploration de ces concepts requiert la constitution de corpus spécialisés permettant d'identifier, de contextualiser et d'analyser les usages lexicaux qui leur sont associés [6]. Ces corpus ne servent pas seulement à enrichir les ressources existantes : ils fournissent une base empirique pour la création de jeux de données destinés à l'analyse linguistique, comparative ou socio-économique. Dans le cadre de cette étude consacrée à « l'innovation non technologique », l'objectif est de construire une ressource multilingue permettant d'observer et de comparer les variations lexicales à travers différents secteurs d'activité.

Ce travail s'inscrit dans le projet MALANTIN, qui développe une approche systématique pour la collecte, le nettoyage et la structuration de données textuelles issues du web et de rapports annuels d'entreprises. Le corpus proposé combine extraction automatisée, détection et vérification de la langue, filtrage des contenus non pertinents et enrichissement en métadonnées. À partir de ce corpus initial, un *dataset* dérivé en anglais est construit pour des usages en apprentissage automatique : il regroupe des blocs contextuels contenant des occurrences de termes annotés selon leur catégorie thématique. L'ensemble du dispositif garantit la traçabilité du processus, la reproductibilité des résultats et la conformité éthique du partage des données.

La section suivante présente les travaux relatifs à ce domaine et situe la méthodologie proposée par rapport aux recherches existantes sur les corpus multilingues et l'analyse de concepts émergents.

# 2. ÉTAT DE L'ART

Les corpus thématiques constituent un outil essentiel pour l'analyse des représentations linguistiques, culturelles et sociales. Au-delà des ressources lexicales générales évoquées dans l'introduction, plusieurs travaux ont exploré la création de corpus spécialisés permettant de documenter la diversité des usages linguistiques dans des environnements variés et de soutenir l'étude de notions complexes ou en évolution.

La montée en puissance des corpus issus du web a notamment joué un rôle structurant. Plusieurs travaux ont montré que l'extraction massive de textes en ligne permettait de constituer des corpus étendus et représentatifs [7], [8]. Dans cette perspective, des études telles que [9] ou [10] ont confirmé que les données web offraient un terrain privilégié pour observer la variabilité lexicale et sémantique dans des contextes réels. De plus, les travaux sur l'identification automatique de la langue ont renforcé la qualité des pipelines de collecte et de prétraitement des données textuelles [11].

Pour aller au-delà des ressources généralistes, plusieurs initiatives récentes se sont orientées vers des corpus thématiques enrichis d'annotations structurées. Le projet MOSAICo [12] relie le texte brut à des bases de connaissances pour permettre une annotation interconnectée. Cette approche s'inscrit dans la continuité des travaux méthodologiques décrits dans le *Handbook of Linguistic Annotation* [13]. D'autres travaux, comme AlphaMWE [14], enrichissent les corpus parallèles par des annotations lexicales fines, portant sur les expressions polylexicales.

---

[1] https://wordnet.princeton.edu/

Enfin, des infrastructures telles que SDHSS [15] qui s'appuie sur des modèles conceptuels FAIR comme CIDOC CRM, ou la plateforme européenne CLARIN [16] renforcent la diffusion, la documentation, l'interopérabilité et la durabilité des données dans des contextes interdisciplinaires.

Ces travaux marquent une évolution vers des corpus sémantiquement enrichis, mieux intégrés dans les écosystèmes de données des SHS. Cependant, la question des concepts émergents, souvent flous, évolutifs et encore dépourvus de formes lexicales stabilisées, demeure largement ouverte. Peu de méthodes existantes proposent un pipeline reproductible permettant de les détecter, de les documenter et de les structurer à partir de corpus multilingues contextualisés.

La méthodologie présentée dans cet article s'inscrit dans cette perspective : elle propose une approche complète, allant de la collecte automatisée à la structuration sémantique, pour construire et exploiter des corpus thématiques destinés à l'identification et à l'étude de concepts émergents.

## 3. DESCRIPTION DE LA MÉTHODOLOGIE

### 3.1. CADRE GÉNÉRAL ET SOURCES DE DONNÉES

La méthodologie adoptée pour la construction du corpus repose sur une approche hybride et systématique, combinant extraction automatisée, filtrage sémantique, structuration multilingue et vectorisation des contenus. Elle vise à constituer une ressource homogène, représentative des contenus textuels décrivant les activités des entreprises, et à offrir un espace d'observation pour la détection des contextes liés à l'innovation. Le corpus repose sur deux types de documents complémentaires : les contenus textuels extraits des sites web et les rapports annuels des entreprises.

Cinq mots-clés conceptuels (*innovation*, *recherche*, *développement*, *stratégie* et *design*) ont servi de guide à l'ensemble du processus, depuis la sélection des sources jusqu'à l'extraction lexicale. Les entreprises ciblées proviennent de la base de données CIB (*Corporate Innovation Base*) du projet européen RISIS 2 (*Research Infrastructure for Science and Innovation Policy Studies*), qui recense les principales firmes européennes impliquées dans des activités d'innovation et de R&D (Recherche et Développement). La base CIB fournit un échantillon d'entreprises impliquées dans des activités d'innovation, sélectionnées selon des critères harmonisés au niveau international. Elle offre un point d'entrée structuré pour identifier des acteurs industriels pertinents dans divers domaines technologiques et économiques.

Au total, 3 992 grandes entreprises, réparties dans 27 secteurs économiques, ont été retenues et validées par les experts du projet. Cette base constitue le jeu de référence pour l'exploration des sites web et la collecte des rapports annuels, garantissant la cohérence sectorielle et linguistique du corpus. Les données collectées sont exclusivement exploitées à des fins de recherche scientifique et transformées en représentations vectorielles pour garantir la conformité éthique et juridique [17] de leur diffusion.

### 3.2. SOUS-CORPUS PDF

Le sous-corpus PDF regroupe les rapports annuels publiés par les entreprises de la base CIB. Ces documents décrivent les orientations stratégiques, les investissements technologiques ainsi que les activités économiques et organisationnelles, ce qui en fait une source adaptée pour analyser la mise en discours de l'innovation dans différents contextes industriels. Ils permettent d'identifier les usages lexicaux liés à l'innovation technologique ou non technologique et d'observer leurs variations selon les secteurs.

La collecte a été réalisée à l'aide d'un analyseur automatisé développé sur la base de la bibliothèque *libcurl* (outil *open* source de gestion de requêtes HTTP), permettant le téléchargement en masse des rapports tout en appliquant un filtrage strict des URL selon des critères prédéfinis, fondés sur des motifs lexicaux et structurels (inclusion du domaine de l'entreprise, exclusion des URL associées à des réseaux sociaux ou à des sites commerciaux ou de divertissement).

Cette procédure repose sur un fichier dictionnaire associant le nom et le domaine de chaque entreprise, fourni par RISIS 2. Chaque entrée de ce fichier, formulée selon la syntaxe :

*COMPANY + intext:"Annual Report" AND inurl:domain*

a servi de requête automatisée via l'API de recherche Google ou par interrogation semi-automatisée, garantissant la localisation des rapports officiels. Les formats de traitement de texte (.doc, .rtf) ont été écartés en raison de leur hétérogénéité structurelle, afin de conserver uniquement les fichiers PDF standardisés et compatibles avec l'extraction automatisée.

Pour compléter la collecte, des rapports supplémentaires ont été récupérés depuis le site *AnnualReports.com* (https://www.annualreports.com), qui centralise les publications financières internationales.

La collecte a été réalisée dans le respect des politiques d'accès des sites (robots.txt) et sans contournement de dispositifs de protection. Seules les ressources librement accessibles ont été traitées conformément aux bonnes pratiques de *scraping* éthique et légal [18].

Les textes extraits ont fait l'objet d'un prétraitement comprenant :

– une normalisation typographique (suppression des caractères accentués et des retours à la ligne, harmonisation des espaces),
– la tokenisation :
– un filtrage lexical minimal destiné à éliminer les segments dépourvus d'information linguistique utile.

Deux seuils de qualité ont été appliqués : exclusion des rapports antérieurs à 2017 et des documents contenant moins de 1 000 tokens après nettoyage, afin de garantir une densité informationnelle suffisante. Une procédure de détection automatique des doublons a permis de ne conserver qu'un rapport par entreprise et par année, en privilégiant le plus récent.

Au total, 8 368 rapports annuels publiés entre 2017 et 2021 ont été rassemblés, assurant une couverture sectorielle et temporelle cohérente avec le périmètre d'étude. Ces documents constituent un sous-corpus homogène et comparable, adaptés à l'analyse textuelle et à l'extraction terminologique. Ils serviront de base à la création de représentations vectorielles et à l'identification des contextes lexicaux présentés dans les sections suivantes.

3.3. SOUS-CORPUS WEB

Le corpus web regroupe les contenus textuels publiés sur les sites des entreprises, décrivant leurs activités, leurs projets et leurs orientations stratégiques. Ces textes permettent d'identifier des segments susceptibles d'évoquer l'innovation, qu'elle soit technologique ou non technologique. Un ensemble de *scrapers Python*, développés à l'aide du framework *Scrapy* [19], a été déployé pour explorer automatiquement les domaines des 3 992 entreprises de la base CIB.

**a. Filtrage et sélection des pages**

Avant toute extraction textuelle, un filtrage systématique des URL a été appliqué afin de limiter le bruit informationnel et de se concentrer sur les pages présentant un contenu pertinent. Des expressions régulières ont permis d'exclure plusieurs catégories de pages non informatives ou non pertinentes pour l'analyse des activités et projets des entreprises :

- pages d'authentification ou de compte utilisateur (*login, register, password reset*) ;
- contenus institutionnels génériques (*FAQ, legal notice, cookie policy*) ;
- sections de recrutement (*careers, jobs, people*) ;
- pages techniques ou utilitaires (*view, print, thumbnail*).

L'extraction a ciblé les balises HTML à forte valeur sémantique : <p>, <title>, <h1> et <h2>. Les pages contenant au moins un des cinq mots-clés conceptuels définis en amont ont été retenues comme points d'entrée pour la sélection des contextes pertinents.

## b. Détection linguistique et enrichissement

La détection de la langue a été effectuée en deux étapes complémentaires. Dans un premier temps, l'information linguistique déclarée dans la balise HTML<lang> a été utilisée lorsqu'elle était disponible. Dans un second temps, une vérification automatique a été réalisée à l'aide de la bibliothèque *googletrans* (https://pypi.org/project/googletrans/), afin de corriger les erreurs fréquentes de métadonnées et d'assurer une identification fiable.

Chaque document a ensuite été enrichi par un ensemble de métadonnées descriptives : nom de l'entreprise, URL, source, secteur d'activité, langue détectée et occurrences de mots-clés conceptuels. Les textes ont été exportés aux formats texte brut et CSV afin de faciliter les traitements et exploitations ultérieurs.

## c. Couverture et diversité linguistique

La collecte a permis d'exploiter 924 sites web d'entreprises, couvrant 27 secteurs économiques et 16 langues principales, après application d'un seuil de qualité fixant à 1 000 le nombre minimal de pages accessibles par domaine. Bien que tous les domaines initialement identifiés n'aient pas pu être intégralement explorés, l'échantillon obtenu reste représentatif du périmètre visé, tant sur le plan sectoriel que linguistique. Les secteurs d'activité couverts sont les suivants :

| Secteurs d'activité |
| --- |
| Machines électriques, électronique industrielle |
| Chimie, pétrole, caoutchouc et plastique |
| Services aux entreprises |
| Fabrication d'équipements de transport |
| Communications |
| Métaux et produits métalliques |
| Logiciels informatiques |
| Commerce de gros |
| Produits alimentaires, fabrication de tabac |
| Banque, assurance et services financiers |
| Biotechnologie et sciences de la vie |
| Matériel informatique |
| Mines et extraction |
| Services publics |
| Produits en cuir, pierre, argile et verre |
| Fabrication de meubles |
| Construction |
| Médias et diffusion |
| Fabrication de textiles et de vêtements |
| Commerce de détail |
| Transport, fret et stockage |
| Fabrication diversifiée |
| Voyages, services personnels et loisirs |
| Administration publique |
| Impression et édition |
| Agriculture, horticulture et élevage |
| Services immobiliers |

*Tableau 1 - Secteurs d'activité des entreprises*

Cette diversité sectorielle et linguistique constitue une base solide pour les étapes de traitement automatisé et de vectorisation décrites dans les sections suivantes.

3.4. PIPELINE TECHNIQUE ET ALGORITHME

Le pipeline technique repose sur une architecture modulaire combinant l'extraction automatisée, la détection linguistique, le filtrage lexical fondé sur les mots-clés conceptuels, la structuration des contenus textuels et, dans une étape ultérieure du pipeline, la production de représentations vectorielles non réversibles. Cette organisation garantit une chaîne de traitement cohérente, reproductible et conforme aux exigences éthiques.

**a. Extraction automatisée et structuration du corpus**

Les documents collectés (pages web et rapports pdf) sont intégrés dans un pipeline automatisé comprenant quatre opérations principales :

- analyse HTML ;
- détection linguistique ;
- filtrage du contenu en fonction de la présence de mots-clés conceptuels;
- extraction et agrégation des données.

Chaque document textuel collecté est intégré dans ce pipeline, analysé et nettoyé, enrichi de métadonnées, puis agrégé dans les fichiers de sortie. Le schéma algorithmique suivant décrit la procédure complète de collecte et de construction du corpus.

*Algorithm 1 – Procédure unifiée de collecte web et construction du corpus*

```
Entrée  : liste de domaines d'entreprises D, lexiques {keywords[lang]}
Sorties : corpus multilingue C, fichiers CSV de synthèse, (optionnel) liste de
liens PDF identifiés
1:  procedure BuildCorpus(D)
2:     C ← ∅
3:     for all domain ∈ D do
4:        assert is_allowed_by_robots(domain)
5:        html_pages ← crawl(domain, rate_limit=polite)
6:        html_pages ← apply_url_filters(html_pages)
7:        for all page ∈ html_pages do
8:           text_raw ← extract_visible_text(page, tags={p,title,h1,h2})
9:           text ← normalize_whitespace(text_raw)
10:          if length_in_tokens(text) < MIN_TOKENS then continue
11:          lang_html ← detect_from_html_attribute(page)
12:          lang_auto ← detect_language(text)
13:          lang ← validate(lang_html, lang_auto)
14:          K ← keywords[lang]
15:          if contains_any(text, K) then
16:             counts ← count_occurrences(text, expand_variants(K, lang))
17:             snippet_set ← extract_context_snippets(text, K, window=W)
18:             snippet_set ← deduplicate_snippets(snippet_set)
19:             meta ← {domain, url=page.url, lang, sector(domain), token_count, date_seen=now()}
20:             C ← C ∪ build_records(snippet_set, counts, meta)
```

```
21:        end if
22:        pdf_links ← find_pdf_links(page)
23:        pdf_links ← apply_url_filters(pdf_links)
24:        for all link ∈ pdf_links do
25:          if is_annual_report(link) then
26:            enqueue_pdf_for_pipeline_3_2(link, domain)
27:          end if
28:        end for
29:      end for
30:    end for
31:    C ← deduplicate_records(C, keys={domain,url,snippet_hash})
32:    aggregate_results_to_csv(C)
33:    return C
34: end procedure
```

Cette architecture modulaire permet une adaptation aisée à d'autres ensembles d'entreprises ou domaines thématiques. Elle garantit également la traçabilité des opérations, la reproductibilité du processus et la conformité éthique de la collecte, notamment vis-à-vis des règles d'accès aux contenus en ligne.

**b. Vectorisation et production des données dérivées**

En complément de la structuration textuelle, une représentation vectorielle non réversible a été produite afin de permettre la diffusion d'un jeu de données exploitable tout en respectant les contraintes légales liées aux droits d'auteur et aux données personnelles. Les phrases anonymisées, obtenues au moyen du pipeline Presidio assurant la reconnaissance automatique d'entités sensibles et leur substitution standardisée, constituent l'entrée du module de vectorisation.

Un tableau de données comprenant quatre champs (word_labels, keywords, source, sentence_anonymized) a servi de base au processus. Les phrases anonymisées ont été encodées à l'aide du modèle paraphrase-multilingual-MiniLM-L12-v2 (Sentence-Transformers), générant pour chaque entrée un vecteur dense de dimension fixe.

Afin d'obtenir des représentations compactes et irréversibles, une réduction dimensionnelle en deux étapes a été appliquée :

- Une Analyse en Composantes Principales (ACP), avec un maximum de 128 composantes, sous la contrainte $n_{composantes} \leq min(n_{échantillons} - 1, n_{caractéristiques})$,
- une projection UMAP en 64 dimensions, adaptée aux usages exploratoires tels que le clustering, la visualisation ou la classification supervisée.

Les représentations finales sont distribuées sous forme de fichiers *sentence_embeddings_pca_128.npy* et *sentence_embeddings_umap_64.npy*, accompagnés de métadonnées non sensibles (*word_labels, keywords, source, doc_id* éventuel). Aucun contenu textuel n'est publié, garantissant la conformité éthique et juridique du *dataset* dérivé.

3.5. FORMALISATION DU PROCESSUS : LA DÉMARCHE PRISME

Afin de garantir la transparence et la reproductibilité du processus, la méthodologie a été structurée selon une approche inspirée du cadre PRISMA (*Preferred Reporting Items for Systematic Reviews and Meta-Analyses*; Page et al., 2021).

Adaptée à la constitution de corpus multilingues en SHS, cette approche est reformulée sous le sigle PRISME, qui désigne six étapes opérationnelles du pipeline : Préparation, Repérage, Identification, Structuration, Modélisation et Évaluation.

Chaque étape correspond à une phase clé du traitement, depuis la définition du périmètre conceptuel jusqu'à la validation du corpus final.

| Étape | Objectif | Procédure | Outils principaux |
|---|---|---|---|
| **P**réparation | Définir le périmètre conceptuel et linguistique du corpus. | Sélection des concepts-clés (*innovation, recherche, développement, stratégie, design*) et constitution de la liste initiale d'entreprises. | Dictionnaire de mots-clés, base CIB (projet RISIS). |
| **R**epérage | Identifier et collecter les sources pertinentes. | *Crawling* sélectif des domaines d'entreprises et recherche de rapports annuels. | Framework *Scrapy*, bibliothèque *libcurl*. |
| **I**dentification | Extraire et segmenter les contenus textuels pertinents. | Analyse du DOM et extraction ciblée des balises `<p>`, `<title>`, `<h1>`, `<h2>`. | XPath, BeautifulSoup. |
| **S**tructuration | Nettoyer et annoter les données textuelles. | Suppression du bruit, détection automatique de la langue, enrichissement par métadonnées. | *googletrans*, *langdetect*, *pandas*. |
| **M**odélisation | Organiser les sous-corpus et générer les fichiers structurés. | Production des sous-corpus web et PDF, exportation en fichiers CSV et texte brut, génération des représentations vectorielles non réversibles. | Scripts Python, pipeline d'analyse interne. |
| **É**valuation | Vérifier la qualité et la couverture du corpus. | Contrôle manuel d'un échantillon, analyse de la distribution lexicale et des occurrences par langue. | Analyse *n-grams*, validation interlinguistique. |

*Tableau 2 - Démarche PRISME pour la constitution du corpus*

Cette formalisation rend le processus de collecte transparent, reproductible et documenté, en précisant les outils, les paramètres et les critères appliqués à chaque étape. Le respect des contraintes d'accès aux sites web (robots.txt, limitations de trafic) assure la conformité éthique de la collecte, tandis que les données dérivées produites sous des formats standardisés (CSV, représentations vectorielles) garantissent la réutilisabilité du corpus pour des tâches d'analyse lexicale ou d'apprentissage automatique.

La section suivante présente l'implémentation concrète de cette méthodologie, depuis le prétraitement linguistique des textes collectés jusqu'à l'extraction et la structuration du lexique thématique.

3.6. R<small>EPÉRAGE ET STRUCTURATION DES TERMES CANDIDATS</small>

Cette section décrit la phase de repérage et d'organisation des termes-clés, conçus comme indicateurs lexicaux permettant d'identifier des contextes potentiellement associés à « l'innovation non technologique ». L'objectif n'est pas de constituer un lexique exhaustif, mais de définir un ensemble structuré de termes de référence permettant de repérer les segments textuels où peuvent émerger de nouvelles formulations liées à la notion d'innovation. Cette phase combine analyse de surface (*n*-grammes) et pondération statistique (TF-IDF), afin d'identifier des termes représentatifs et fréquemment cooccurrents dans les textes d'entreprises.

**a. Première étape : génération des *n*-grammes**

Les textes collectés sont d'abord segmentés en *n*-grammes (séquences de *n* mots), avec *n* variant entre 2 et 5. Cette approche, indépendante des contraintes linguistiques spécifiques, permet de repérer des associations lexicales récurrentes y compris dans des langues et styles rédactionnels variés.

Un premier filtrage a été effectué pour écarter les *n*-grammes contenant des noms d'entreprises ou des séquences non lexicales (liens URL, adresses, éléments numériques, etc.).

Afin de concentrer l'analyse sur les domaines d'intérêt, seuls les *n*-grammes comportant au moins un mot appartenant à un ensemble de mots-cibles ont été conservés. Ces mots-cibles corres-pondent aux cinq concepts directeurs du projet (innovation, recherche, développement, stratégie, design) enrichis de deux termes additionnels (*new*, *product*), fréquemment associés aux contextes d'innovation selon les experts.

Enfin, un filtrage statistique a été appliqué afin de ne conserver que les *n*-grammes les plus significatifs, c'est-à-dire ceux dont la fréquence se situait au-delà de trois écarts types au-dessus de la médiane de distribution.

**b. Deuxième étape : pondération *TF-IDF***

La seconde étape consiste à appliquer la pondération *TF-IDF* aux *n*-grammes issus du premier filtrage. Cette mesure statistique évalue l'importance d'un *n*-gramme dans un domaine d'entreprise au regard de sa fréquence dans l'ensemble du corpus. Chaque texte d'entreprise est traité comme un document unique, regroupant les contenus extraits des différentes URL d'un même domaine. Les textes sont ensuite comparés au sein de leur secteur d'activité, ce qui permet d'identifier les termes spécifiques à chaque domaine économique.

L'objectif est de faire émerger un lexique de référence fondé sur les régularités lexicales observées, plutôt que de partir d'un vocabulaire préétabli.

**c. Illustration intersectorielle : pondérations comparées**

Le tableau 3 présente, à titre d'exemple, les termes les mieux pondérés (selon la méthode TF-IDF) dans deux secteurs distincts, ainsi que leur intersection. Les valeurs indiquent le poids relatif de chaque terme dans l'ensemble des documents analysés, ce qui permet d'identifier les dénominations les plus caractéristiques du lexique de l'innovation dans chaque domaine.

Certains *n*-grammes, tels que *new product* ou *product development*, apparaissent plusieurs fois dans les deux secteurs en raison de leur fréquence élevée dans des contextes sémantiquement distincts.

| Services aux entreprises | Poids du terme | Machines électriques et électroniques industrielles | Poids du terme | Intersection |
|---|---|---|---|---|
| research development | 1,00 | research development | 1,00 | research development |
| sustainable development | 0,73 | products services | 0,71 | sustainable development |
| product development | 0,65 | product development | 0,69 | product development |
| product services | 0,57 | new product | 0,45 | product services |
| business development | 0,54 | sustainable development | 0,43 | business development |
| new product | 0,49 | business development | 0,42 | new product |
| innovative solution | 0,46 | innovative product | 0,40 | innovative solution |
| new technology | 0,41 | design development | 0,38 | new technology |
| new product | 0,40 | new product | 0,38 | new product |
| software development | 0,37 | innovative solution | 0,37 | innovative product |
| product solution | 0,36 | new technology | 0,36 | —[2] |
| innovative product | 0,35 | clinical research | 0,32 | — |
| new business | 0,35 | product design | 0,29 | — |
| development process | 0,35 | development manufacturing | 0,29 | — |
| innovation award | 0,34 | new idea | 0,26 | — |
| development goal | 0,34 | quality product | 0,25 | — |

***Tableau 3:** Intersection des termes les mieux pondérés par secteur*

---

[2] Absence du terme dans l'intersection

Un contrôle de redondance a également été effectué afin d'éliminer les paragraphes dupliqués, fréquents sur les sites web institutionnels. Les *n*-grammes présentant un score TF-IDF supérieur à un seuil défini sur la distribution des valeurs observées ont été retenus comme termes candidats pour le lexique final.

Cette double approche, linguistique (*n*-grammes) et statistique (TF-IDF), permet d'obtenir une sélection robuste de termes pertinents pour l'analyse thématique et la constitution de ressources lexicales sectorielles. Elle offre également une base exploitable pour l'apprentissage automatique, en facilitant la création de jeux de données annotés par secteur ou par domaine.

La section suivante présente les résultats issus de ce processus, en mettant en évidence la taille du corpus, la répartition sectorielle et linguistique ainsi que les premiers indices contextuels liés à l'innovation non technologique.

## 4. RÉSULTATS

Le corpus final couvre une grande variété de secteurs économiques et de langues, offrant une richesse lexicale autour des cinq mots-clés conceptuels. Les résultats présentés ici portent sur la constitution et la structuration du corpus, et non sur les modèles d'apprentissage automatique dérivés. L'analyse des données collectées montre une couverture sectorielle étendue, ainsi qu'une densité lexicale suffisante pour envisager des tâches ultérieures d'extraction ou de modélisation thématique.

| Langue | Nombre d'URL | Secteurs couverts | Nombre de tokens |
|---|---|---|---|
| Anglais | 1 224 | 27 | 8 133 370 |
| Chinois | 3 054 | 22 | 1 823 235 |
| Français | 5 074 | 17 | 2 821 719 |
| Allemand | 6 783 | 16 | 3 327 026 |
| Espagnol | 0 364 | 12 | 2 483 900 |
| Italien | 236 | 12 | 1 323 276 |
| Coréen | 0 929 | 11 | 1 899 763 |
| Néerlandais | 907 | 11 | 505 163 |
| Catalan | 427 | 9 | 1 149 180 |
| Portugais | 594 | 8 | 780 462 |
| Danois | 088 | 8 | 234 798 |
| Russe | 2 088 | 7 | 13 157 829 |
| Suédois | 355 | 7 | 219 159 |
| Polonais | 632 | 6 | 241 973 |
| Finnois | 029 | 4 | 99 361 |
| Ukrainien | 073 | 1 | 981 |

*Tableau 4 - Répartition linguistique, sectorielle et volumétrique du corpus*

La diversité linguistique observée reflète la variété lexicale et thématique attendue dans un corpus de grande ampleur couvrant plusieurs secteurs. Cette variété assure une meilleure représentativité des usages terminologiques et renforce la robustesse de la ressource pour des analyses comparatives et des expérimentations en traitement automatique du langage.

Le corpus ainsi obtenu constitue une ressource multilingue structurée, prête à être exploitée pour des analyses lexicométriques, l'extraction terminologique ou la création d'ensembles de données destinés à l'apprentissage supervisé. Dans cette perspective, un jeu de données dérivé sous forme des représentations vectorielles non réversibles a été produit, spécifiquement conçu pour les tâches d'apprentissage automatique.

Ce jeu de données est structuré autour d'un lexique de référence organisé en sept catégories thématiques (non détaillées ici), regroupant les termes potentiellement liés à l'innovation. Pour chaque terme du lexique, des blocs de contexte d'environ 500 tokens avant et après l'occurrence ont été extraits du corpus principal, constituant des exemples d'apprentissage supervisé. L'entraînement du modèle sur ces segments a permis de proposer de nouveaux termes associés à l'innovation non technologique.

Après vérification par les experts du projet, plus de 80 % des termes suggérés (dans un total de 700 termes proposés) ont été jugés pertinents et intégrés au lexique final, démontrant la fiabilité et l'efficacité du processus hybride combinant analyse automatique et validation humaine.

## 5. CONCLUSION

En réponse aux besoins de construction de corpus destinés à l'identification de concepts émergents, la méthodologie proposée démontre la faisabilité d'une approche hybride, reproductible et évolutive pour la constitution de corpus multilingues thématiques.

Le dispositif associe extraction automatisée, structuration lexicale et annotation statistique, afin de rendre compte des régularités lexicales observées dans des textes issus de sources hétérogènes, tout en garantissant leur traçabilité.

Le jeu de données dérivé, intégralement vectorisé, constitue une ressource réutilisable et interopérable pour la recherche et l'apprentissage automatique. Il regroupe les blocs contextuels extraits autour des termes-clés identifiés, enrichis de métadonnées descriptives (secteur, langue, score TF-IDF, *n*-gramme) et d'un encodage vectoriel assurant la conformité éthique et juridique du partage. Ce format de diffusion préserve la valeur scientifique du corpus tout en éliminant les risques liés à la redistribution de contenus protégés.

Le corpus agrégé (entraînement et test) est disponible en ligne[3], sous licence ouverte pour la recherche, dans un format compatible avec les tâches de classification, de détection lexicale et de modélisation sémantique.

L'intégration d'un apprentissage supervisé, combinée à une validation experte, a permis d'identifier et de confirmer de nouveaux termes liés à l'innovation non technologique, renforçant la robustesse du protocole et sa portée interdisciplinaire. Cette approche offre une passerelle entre méthodes computationnelles et analyse linguistique, en rendant possible l'exploration systématique de dynamiques lexicales dans des champs conceptuels en évolution.

Au-delà du cas d'étude présenté, ce travail contribue à combler le manque de ressources adaptées aux besoins des sciences humaines et sociales, en facilitant leur réutilisation dans des cadres ouverts et collaboratifs.

Les perspectives incluent l'extension du protocole à d'autres domaines conceptuels, la normalisation des métadonnées selon les principes FAIR, et l'intégration progressive de ces corpus dans des infrastructures partagées telles que SDHSS (*Semantic Data for Humanities and Social Sciences*).

Ainsi, le projet illustre la complémentarité entre l'informatique et les sciences humaines, en croisant ingénierie linguistique et analyse des dynamiques d'innovation à partir de corpus textuels d'entreprises.

## 6. REMERCIEMENTS

---

[3] https://repository.ortolang.fr/api/content/malantin/head/
https://entrepot.recherche.data.gouv.fr/dataset.xhtml?persistentId=doi:10.57745/M9UHMX&version=DRAFT

ses remarques constructives sur la présentation méthodologique. Elles expriment enfin leur gratitude à Julien Muller (CNRS, comité de rédaction de Data & Corpus) pour ses conseils et suggestions éditoriales lors de la préparation de cet article, ainsi qu'aux stagiaires et partenaires du projet.

## 7. CONTRIBUTION

Anna Pappa : conceptualisation, méthodologie, ressources, développement, supervision, rédaction - ébauche du manuscrit, relectures et corrections.

Revekka Kyriakoglou : analyse formelle, ressources, développement, tests, validation, rédaction, relectures et corrections.

## 8. RÉFÉRENCES BIBLIOGRAPHIQUES SCIENTIFIQUES